\documentclass[a4paper, 10pt, conference]{ieeeconf}      
   \usepackage[table,xcdraw]{xcolor}                                                       
\usepackage{FG2025}

\FGfinalcopy 

\IEEEoverridecommandlockouts                              
\usepackage{graphics} 
\usepackage{epsfig} 
\usepackage{mathptmx} 
\usepackage{times} 
\usepackage{amsmath} 
\usepackage{amssymb}  
\usepackage{booktabs}
\usepackage{multicol,multirow}
\usepackage{adjustbox}
\usepackage{url}
\usepackage{subcaption}

\def\FGPaperID{156} 

\title{\LARGE \bf
Are Foundation Models All You Need for Zero-shot Face Presentation Attack Detection?
}

\author{\parbox{16cm}{\centering
    {\large Lazaro Janier Gonzalez-Soler and Juan E. Tapia and Christoph Busch}\\
    {\normalsize
    da/sec - Biometrics and Security Research Group, Darmstadt, Germany}} \\
    \normalsize \{lazaro-janier.gonzalez-soler;juan.tapia-farias;christoph-busch\}@h-da.de
    \thanks{This research work has been partially funded by the European Union (EU) under G.A. no. 101121280 (EINSTEIN) and CarMen (101168325), and UKRI Funding Service under IFS reference 10093453, and the German Federal Ministry of Education and Research and the Hessian Ministry of Higher Education, Research, Science and the Arts within their joint support of the National Research Center for Applied Cybersecurity ATHENE.}
}

\usepackage{fancyhdr}
\begin{document}

\ifFGfinal
\thispagestyle{empty}
\pagestyle{empty}
\else
\author{Anonymous FG2025 submission\\ Paper ID \FGPaperID \\}
\pagestyle{plain}
\fi
\maketitle
\thispagestyle{fancy}

\begin{abstract}

Although face recognition systems have undergone an impressive evolution in the last decade, these technologies are vulnerable to attack presentations (AP). These attacks are mostly easy to create and, by executing them against the system's capture device, the malicious actor can impersonate an authorised subject and thus gain access to the latter's information (e.g., financial transactions). To protect facial recognition schemes against presentation attacks, state-of-the-art deep learning presentation attack detection (PAD) approaches require a large amount of data to produce reliable detection performances and even then, they decrease their performance for unknown presentation attack instruments (PAI) or database (information not seen during training), i.e. they lack generalisability. To mitigate the above problems, this paper focuses on zero-shot PAD. To do so, we first assess the effectiveness and generalisability of foundation models in established and challenging experimental scenarios and then propose a simple but effective framework for zero-shot PAD. Experimental results show that these models are able to achieve performance in difficult scenarios with minimal effort of the more advanced PAD mechanisms, whose weights were optimised mainly with training sets that included APs and bona fide presentations. The top-performing foundation model outperforms by a margin the best from the state of the art observed with the leaving-one-out protocol on the SiW-Mv2 database, which contains challenging unknown 2D and 3D attacks.\footnote{\url{https://github.com/ljsoler/zero-shot-FoundationPAD}}                   

\end{abstract}

\section{INTRODUCTION}
\label{sec:intro}

The development and evolution of face recognition systems over the years has been mainly due to the success of advances in the area of deep learning~\cite{Deng-ArcFace-CVPR-2019,Meng-MagFace-CVPR-2021,Kim-AdaFace-CVPR-2022,George-EdgeFace-TBIOM-2024}. Despite their advances, facial recognition technologies are vulnerable to attack presentations (AP) which, in most cases, can be easily created by a malicious individual with the intent to impersonate an authorised subject and gain access to the latter's information (e.g. financial transactions and unlocking of smartphones). The daily information flow through social networks such as Facebook, Instagram and YouTube allows an attacker to download a photo or video of a target subject and replay it on the system's capture device (this is a 2D attack) to grant unauthorised access to different applications~\cite{Pasmino-FlickrPAD-IWBF-2023}. More sophisticated attacks, including 3D masks, can also be used effectively to circumvent biometric recognition technologies.   

To protect face recognition systems against APs, numerous presentation attack detection (PAD) approaches have been proposed~\cite{Shaheed-PADSurvey-2024}. Current state-of-the-art PAD algorithms are mainly developed upon deep learning and require a large amount of data for training to obtain reliable detection performance~\cite{GonzalezSoler-FaceRegions-2023,Fang-LMFD-WCACV-2022,Fang-CF-PAD-WCACV-2024,Ozgur-FoundPAD-ArXiv-2025}. Despite the progress achieved over the years, these PAD algorithms lack generalisability, which is evidenced by the degradation of their performance in detecting unknown presentation attack instruments (PAI) or databases that have not been seen during training. Note that the collection of new databases to train PAD subsystems has not experienced the same advances as PAD technologies and is partly due to privacy concerns and the fact that it is a time-consuming task. To alleviate the lack of generalisability, the literature has focused, on the one hand, on the creation of synthetic data that resembles real images captured from a PAI~\cite{Fang-Synthaspoof-CVPR-2023,Fang-SynFacePAD-IJCB-2023}. On the other hand, reusing the weights of deep neural networks (DNN) that were optimised with a huge amount of images~\cite{Liu-DeeptreeLearning-CVPR-2019,George-VitEffPAD-IJCB-2021} and they are supposed to be generalisable to different tasks. 

Human learning is inherently multimodal, as harnessing multiple senses together helps us to better understand and analyse new information. Recent advances in multimodal learning have been inspired by the effectiveness of this process in creating models capable of processing and relating information using a variety of modalities such as image, video, text, audio, body gestures, facial expressions and physiological signals. In this paper, we focus in particular on the reuse of DNN weights to mitigate the lack of generalisability of PAD approaches. To do so, we explore the effectiveness of recent foundation models for zero-shot PAD. Foundation models are large models pre-trained on large amounts of data, designed to be generalisable and easily adaptable to specific tasks. Zero-shot classification is the task of predicting objects of unseen classes (target domain) by transferring knowledge obtained from other seen classes (source domain) with the help of semantic information~\cite{Pourpanah-ZeroShot-TPAMI-2022}. Exploiting the generalisable weights of the foundational models, we attempt to provide a simple framework that is capable of detecting unknown PAI with high performance. The main contributions of this work are summarised below:

\begin{itemize}
    \item Demonstration of the effectiveness of the foundation model-based framework on an unrelated top-down task, adapting only a minimum number of parameters related to the classification header in the training phase. It is shown that the performance of the framework for zero-shot PAD is improved by simply fusing different foundation models. 
    

    \item Extensive evaluation in line with metrics defined in the international standard ISO/IEC 30107-3~\cite{ISO-IEC-30107-3-PAD-metrics-2023} for biometric PAD of the proposed approach in challenging scenarios, such as unknown PAI species and cross-database. Experimental evaluation shows that the proposed framework can achieve state-of-the-art performance in different protocols and outperforms baselines by a large margin.
\end{itemize}

 The remainder of this paper is organised as follows: Related work is summarised in Sect.~\ref{sec:related_work}. In Sect.~\ref{sec:framework}, we describe the foundation models-based framework. The experimental setup is summarised in Sect.~\ref{sec:exp_setup}. Experimental results, including the foundation model assessment, as well as a benchmark of the proposed PAD framework on challenging settings, are presented in Sect~\ref{sec:results}. Conclusions and future work directions are finally summarised in Sect.~\ref{sec:conclusions}.

\section{RELATED WORK}
\label{sec:related_work}

To mitigate the threats posed by attacks and thus increase the security of biometric face recognition systems, numerous PAD approaches have been progressively proposed over the last decade. They can be hardware- and software-based~\cite{El-PAD-CNNSurvey-IET-Biometrics-2020,Galbally-PAD-FaceSurvey-2014,Raghavendra-FacePAD-Survey-2017}. With the introduction and success of DNNs, most software-based PAD methods evolved from handcrafted feature analysis~\cite{Arashloo-PAD-BSIFfusion-2015,Gonzalez-PAD-FVencForFacePAD-BIOSIG-2020,GonzalezSoler-UnkownAttacksFace-IET-Biometrics-2021,Raghavendra-PAD-MS-LPQ-2018} to the development of sophisticated convolutional neural networks (CNNs)~\cite{Fang-PatchWise-IJCB-2022,Fang-CF-PAD-WCACV-2024,George-PAD-DeepPixelBis-ICB-2019}, and vision transformers~\cite{George-VitEffPAD-IJCB-2021,Ozgur-FoundPAD-ArXiv-2025}.  

In 2014, Yang \textit{et al.}~\cite{Yang-PAD-CNNApplicability-ArXiv-2014} fine-tuned ImageNet pre-trained CaffeNet~\cite{Jia-PAD-CaffeNet-2014} and VGG-face~\cite{Parkhi-VGG-Face-2015} models for PAD. Based on this idea, Xu~\textit{et al.}~\cite{Xu-PAD-LSTM-2015} combined Long Short-Term Memory (LSTM) units with CNNs to learn temporal features from face videos. Sanghvi \textit{et al.}~\cite{Sanghvi-MixNet-ICPR-2021} enhanced generalisability by combining three CNN sub-architectures, one for each common PAI species, i.e. print, replay and mask attacks. Fang \textit{et al.}~\cite{Fang-LMFD-WCACV-2022} proposed a hierarchical attention module integration to merge information from two streams at different stages, considering the nature of deep features in different layers of the CNN. Some techniques~\cite{Chen-DualStream-3dMask-CVPR-2021,Liu-ContextAware-TIFS-2022} have also proposed CNNs to analyse properties in 3D mask attacks based on the fact that 2D face PAD algorithms suffer from a significant degradation of detection performance in this type of PAI species. Since acquisition properties such as facial appearance, pose, lighting, capture devices, PAI species and even subjects vary between datasets, several major facial PAD approaches have recently explored domain adaptation (DA) to align features from two different domains~\cite{Fang-CF-PAD-WCACV-2024,Li-KnownledgeDistillation-TIFS-2022,deFreitas-FaceDomainAdaptation-2019,Wang-PAD-AdvDomainAdaptation-TIFS-2020,Wang-MDIL-AI-2024,Yang-PAD-PersonSpecDA-TIFS-2015}. 

While PAD approaches have achieved good results in unseen target domains, they depend on the availability of labelled data from various sources, which is difficult to satisfy in practice.  Due to privacy concerns in biometric data acquisition, PAD algorithms are trained on small databases containing a limited number of domains, resulting in a lack of generalisability~\cite{Ozgur-FoundPAD-ArXiv-2025}. 


One solution to deal with low data availability in PAD relies on the use of foundation models. These models contain a large number of parameters and are trained on large and diverse datasets, resulting in highly generalisable models that are easily adaptable to different computer vision and pattern recognition tasks~\cite{Awais-FM-TPAMI-2025}. 

Since 2021, there has been an emerging interest in foundational models that combine vision and language modalities (also called joint vision-language models). In 2023, Meta AI presented a vision-language model called SegmentAnything (SAM)~\cite{Kirillov-SAM-ICCV-2023,Ravi-SAM2-ArXiv-2024} which learned the general encoding of any object to achieve zero-shot generalisation to unknown objects and images without requiring additional training. Building upon self-supervised strategies, OpenAI presented Contrastive Language-Image Pretraining (CLIP)~\cite{Radford-CLIP-ICML-2021} that combines text prompts with image encoding through cross-attention mechanisms to learn visual concepts, enabling zero-shot transfer of the model to subsequent tasks. Following this idea, Meta AI introduced DINO~\cite{Oquab-DINOv2-ArXiv-2023}, which can generate universal features for image-level and pixel-level tasks. Google also proposed Large-scale ImaGe and Noisy-Text Embedding (ALIGN)~\cite{Jia-ALIGN-ICML-2021}, which is trained similarly to CLIP (i.e. using contrastive learning~\cite{Wu-ComntrastiveLearning-CVPR-2018} between text-image pairs) to learn a general representation that can be used in subsequent visual and vision-language tasks.

\section{FOUNDATION MODELS-BASED PAD FRAMEWORK}
\label{sec:framework}

\begin{figure}[!tb]
    \centering
    \includegraphics[width=\linewidth]{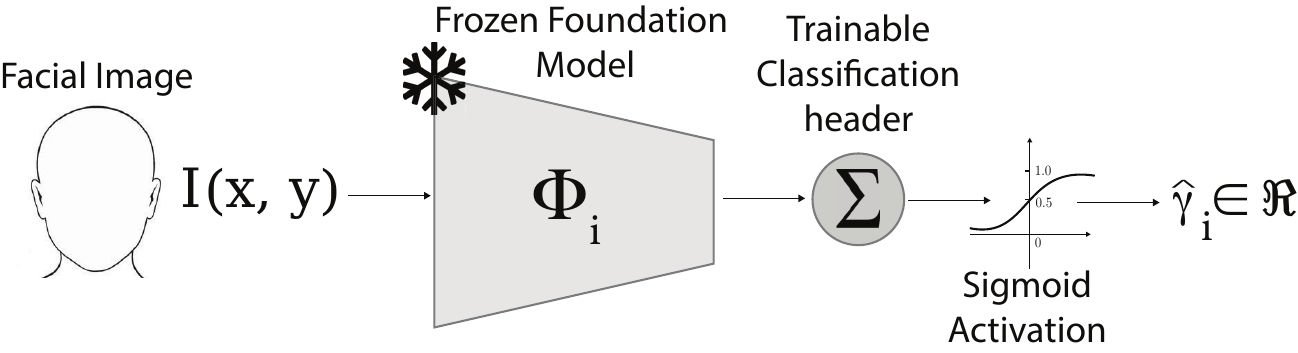}
    \caption{Overview of the foundation model-based framework for zero-shot PAD.}
    \label{fig:overview}
\end{figure}

Despite the increasing attention that foundation models have received in recent years, their application in the field of biometrics remains largely unexplored. To date, a limited number of works focusing on foundation models have addressed facial PAD~\cite{Fang-UnifiedAttacks-ArXiv-2024,Ozgur-FoundPAD-ArXiv-2025}. Most are trained from scratch~\cite{Fang-UnifiedAttacks-ArXiv-2024} or their pre-trained weights are partially optimised~\cite{Ozgur-FoundPAD-ArXiv-2025}, and the extent to which the representation learned by the foundation models can be used for zero-shot PAD remains unexplored. In our work, we investigate the extent to which the pre-trained weights of the foundation models for facial PAD are generalisable. To that end, the combination of highly-performing foundation models is selected and adapted for zero-shot PAD as shown in Fig.~\ref{fig:overview}. 

Consider $I(x,y)$ as the input cropped facial image and $\Phi_i$ a foundation model whose pre-trained weights are frozen, i.e., the pre-trained weights are not altered either during optimisation or inference. The classification header for $\Phi_i$, which consists of the number of classes, is set to a single neuron for the bona fide presentation (BP) vs. AP decision and this will only be optimised during training utilising binary cross-entropy loss, while the remaining weights of the model will remain unchanged. The binary cross-entropy loss is defined for the prediction $\hat{\gamma}_i$ and the respective groundtruth $\gamma_i$ as:   


\begin{equation}
    \mathcal{L} = -(\gamma_i\cdot \mathrm{log} \hat{\gamma}_i + (1 - \gamma_i)\cdot\mathrm{log}(1 - \hat{\gamma}_i))
\end{equation}

In the experiments (Sect.~\ref{sec:results}), we evaluated $N=6$ foundation models. Therefore, $I(x,y)$ runs through $\Phi_i: i \in \{1\ldots N\}$ to computing $\hat{\gamma}_i$. Each $\Phi_i: i \in \{1\ldots N\}$ is first tested for zero-shot PAD and different score-level fusions between $\hat{\gamma}_i: i \in \{1\ldots N\}$ are also evaluated to obtain the final decision $\hat{\gamma}$. To test the extent to which the weights of the foundation models $\Phi_i: i \in \{1\ldots N\}$ are generalisable and can be easily adapted to unrelated top-down tasks such as PAD, several score-level fusions are selected~\cite{Ross-ScoreFusion-BiometricsEnc-2009}.

Let $F \in \{\mathrm{MIN}, \mathrm{MAX}, \mathrm{SUM}, \mathrm{AVG}\}$. Then:
\begin{equation}
    \hat{\gamma} = F(\hat{\gamma}_1, \ldots, \hat{\gamma}_N)
\end{equation}

Note that these fusion strategies are agnostic to the input parameters and do not require a development set for optimisation.  

In our work, we selected two families of different high-performance foundation models that have reported competitive results in zero-shot learning scenarios and a high generalisability across a wide range of tasks~\cite{Radford-CLIP-ICML-2021}: CLIP~\cite{Radford-CLIP-ICML-2021} and DINO~\cite{Oquab-DINOv2-ArXiv-2023}. In contrast to other works~\cite{Ozgur-FoundPAD-ArXiv-2025}, we only use the CLIP image encoder, as the use of the text prompt results in poor performance, as in~\cite{Ozgur-FoundPAD-ArXiv-2025}. We believe that CLIP's pre-trained weights were mostly not optimised with text prompts containing terms such as ``attack presentation'', ``spoofing'', ``bona fide presentation'' and ``real''. Therefore, poor detection performance is to be expected, such as in~\cite{Ozgur-FoundPAD-ArXiv-2025}.            

\begin{table}[!t]
	\centering
	\caption{A summary of databases used in our experiments.}
	\label{tab:DB}
    \begin{adjustbox}{width=\linewidth}
    \addtolength{\tabcolsep}{-0.5em}
	\begin{tabular}{r c r c c l} \toprule
            DB              &   \#Videos              &     Split     &     \#BP    &  \#AP      &  PAI species\\
\midrule
\multirow{2}{*}{CASIA-FASD (C)} & \multirow{2}{*}{600} 	  &     Train     &     60      &  180       &	\multirow{2}{*}{\shortstack[l]{Warped photo (Printed attack), \\ Cut photo, Video replay}}   \\ 
	                	& 						  &	    Test      &     90      &  270       &           \\ 
\midrule
\multirow{3}{*}{REPLAY-ATTACK (I)} & \multirow{3}{*}{1,200}          &     Train  &     60      &  300    & \multirow{3}{*}{\shortstack[l]{Printed attacks, Photo replay, \\ Video replay}} \\
                               &			           &     Dev    &     60      &  300    &               \\ 
                               &			           &     Test   &     80      &  400    &               \\ 
\midrule
\multirow{3}{*}{OULU-NPU (O)} & \multirow{3}{*}{4,950}    &     Train  &     360     &  1,440  & \multirow{3}{*}{\shortstack[l]{Printed attacks, Video replay}} \\
	                   &                           &     Dev    &     270     &  1,080  &                                                  \\
                          &                           &     Test   &     360     &  1,440  &                                                  \\
\midrule
\multirow{2}{*}{MSU-FASD (M)} 	& \multirow{2}{*}{440} 	  &     Train  &     30      &  90     & \multirow{2}{*}{\shortstack[l]{Printed attacks, \\ Video replay}}  \\
						  &					        &     Test   &     40      &  120    &                                            \\
\midrule
\multirow{7}{*}{SiW-Mv2}    & \multirow{7}{*}{1,700}  &\cellcolor{gray!15}  & \multirow{7}{*}{785}& \multirow{7}{*}{915} & \multirow{7}{*}{\shortstack[l]{Funny Eyes (FunE.), Partial Eyes (PEye), \\ Partial Mouth (PMouth), Paper Glasses (PaperG), \\ Obfuscation (Ob.), Impersonation (Impers.), \\ Cosmetic, Half Masks (HalfM.), Silicone, \\ Transparent Masks (TransM.), Paper, \\ Mannequin (Mann.), Video replay, Printed attacks}} \\
						& 						  &  \cellcolor{gray!15} &             &        &  \\
						& 					      &  \cellcolor{gray!15} &             &        &  \\
						& 						  &  \cellcolor{gray!15} &             &        & \\
                            & 						  &  \cellcolor{gray!15} &             &        &  \\ 	
                            & 						  &  \cellcolor{gray!15} &             &        &  \\ 	
                            & 						  &  \cellcolor{gray!15} &             &        &  \\ 	
\bottomrule
	 \end{tabular}
     \end{adjustbox}
\end{table}

\section{EXPERIMENTAL SETUP}
\label{sec:exp_setup}

The main goals of the experimental evaluation are $i)$ to assess the generalisability of foundation models in different operational scenarios for zero-shot PAD and $ii)$ to check to what extent the fusion of these foundation models based on the proposed framework can improve the particular performance reported by each foundation model. The operational scenarios are defined as follows:

\begin{itemize}
    \item \textbf{Known‐attacks} scenario reports an analysis of all PAI species. The scenario is assumed naïve as all PAI species for the test are included in the training set. On this scheme, a benchmark is performed between traditional CNNs (e.g., ResNet~\cite{He-ResNet-CVPR-2016}, DenseNet~\cite{Huang-DenseNet-CVPR-2017}, MobileNet~\cite{Koonce-Mobilenetv3-2021}, and EfficientNet~\cite{Tan-EfficientNetv2-ICML-2021}) and foundation models for zero-shot PAD following the protocol in CASIA-FASD~\cite{Zhang-CASIAFASD-ICB-2012}. 

    \item \textbf{Unknown PAI species} scenario, in which the PAI species used for testing are not incorporated in the training set. We follow the ‘leave-one-out’ test protocol explained in SiW-Mv2~\cite{Guo-SiWMv2-ECCV-2022} in which one PAI species is evaluated at a time while the rest of the PAI species are for training.

    \item \textbf{Cross‐database} is considered the most challenging and realistic as the datasets used for testing are different from those used for training the algorithms. To avoid biases related to external variables, the PAI species for the tests are also included in the training set. To compare the foundation models with the state-of-the-art, cross-database settings widely used in different benchmarks~\cite{Fang-PatchWise-IJCB-2022,Fang-LMFD-WCACV-2022,Fang-CF-PAD-WCACV-2024} are evaluated. 
\end{itemize}

\subsection{Databases}

\begin{figure}[!t]
    \centering
    \begin{tabular}{cc}
        \begin{subfigure}{0.54\linewidth}
            \includegraphics[width=\linewidth]{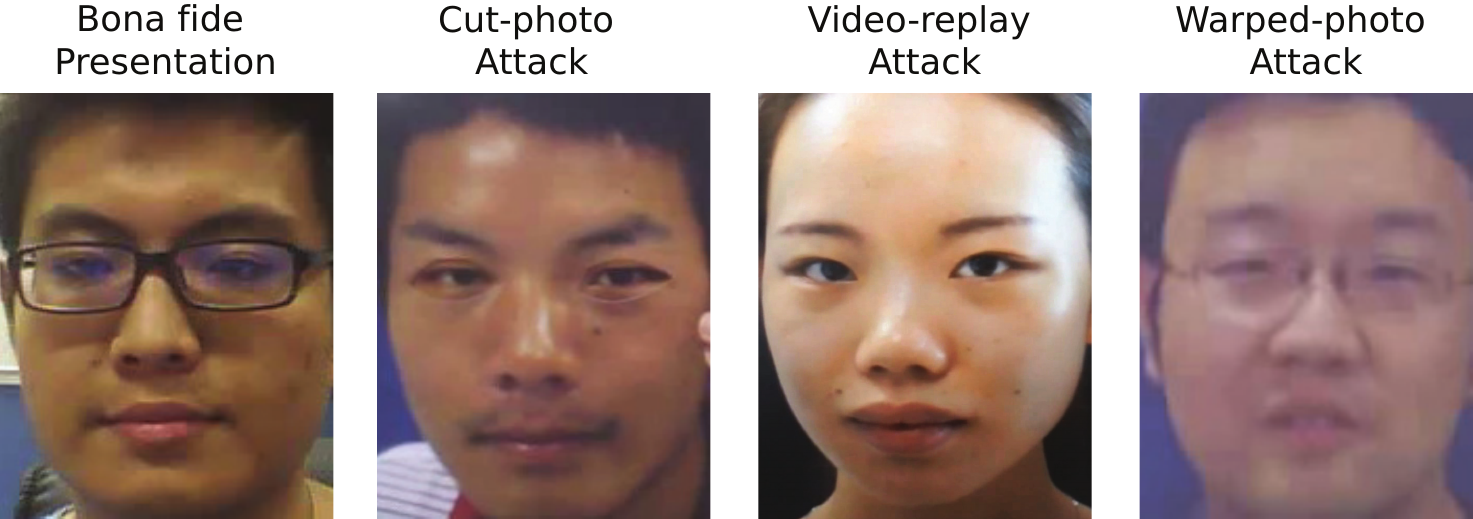}
            \caption{CASIA-FASD}
        \end{subfigure} & 
        \multirow{4}{*}{ 
            \begin{minipage}{0.234\linewidth} 
                \centering
                \vspace{-1.86cm}
                \begin{subfigure}{\linewidth}
                    \includegraphics[width=\linewidth]{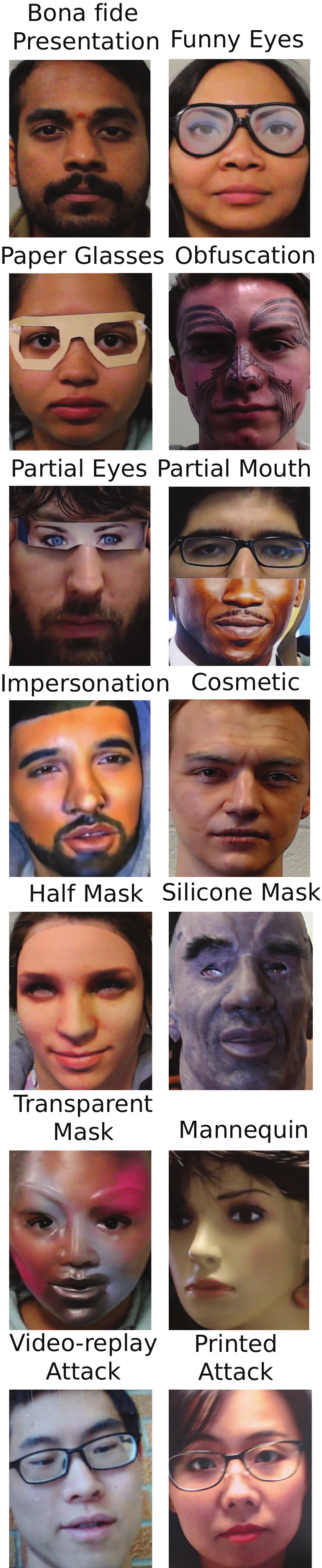}
                    \caption{SiW-Mv2}
                \end{subfigure}
            \end{minipage}
        } \\
        \begin{subfigure}{0.54\linewidth}
            \includegraphics[width=\linewidth]{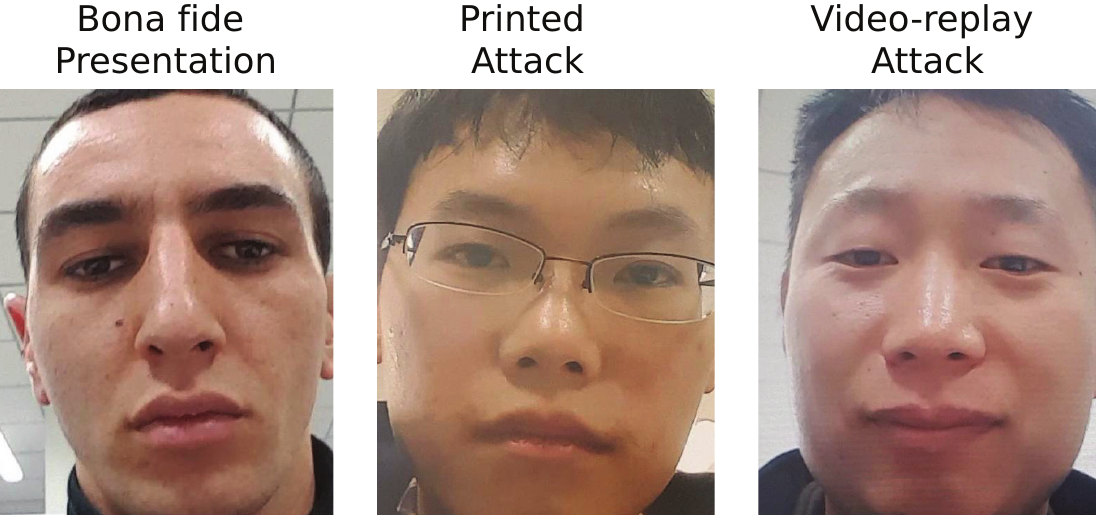}
            \caption{OULU-NPU}
        \end{subfigure} & \\ 
        \begin{subfigure}{0.54\linewidth}
            \includegraphics[width=\linewidth]{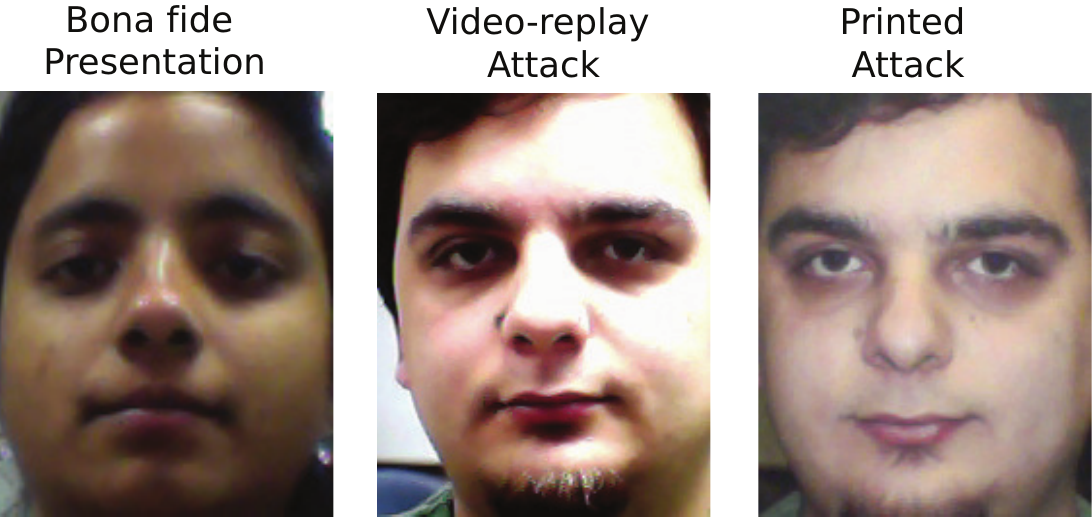}
            \caption{MSU-FASD}
        \end{subfigure} & \\ 
        \begin{subfigure}{0.54\linewidth}
            \includegraphics[width=\linewidth]{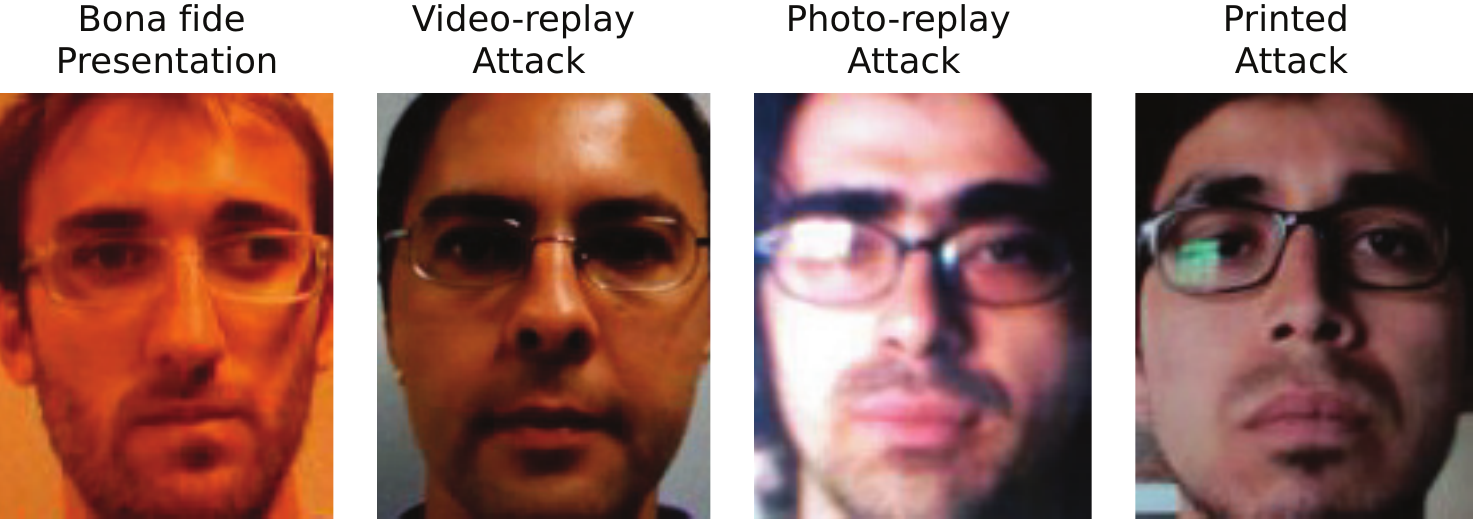}
        \caption{REPLAY-ATTACK}
    \end{subfigure} & \\
    \end{tabular}
    \caption{Example of BP and PAIs in each database used in the experimental evaluation.}
    \label{fig:databases}
\end{figure}

\begin{table*}[!tb]
	\centering
	\caption{Detection performance (in \%) for different foundation models for the known-attack protocol in CASIA-FASD. }
	\label{tab:KA_CASIA}
	\begin{adjustbox}{width=\textwidth}
        \addtolength{\tabcolsep}{-0.4em}
	\begin{tabular}{r| c c c c| c c c c| c c c c| c c c c}  \toprule
        \multirow{3}{*}{\textbf{Approaches}} & \multicolumn{16}{c}{\textbf{PAI species}}  \\
                                        & \multicolumn{4}{c|}{cut-photo attack} & \multicolumn{4}{c|}{video-photo attack} & \multicolumn{4}{c|}{warped-photo attack} & \multicolumn{4}{c}{overall} \\ 
                                        & D-EER  & BPCER10 & BPCER20 & BPCER100 & D-EER  & BPCER10 & BPCER20 & BPCER100 & D-EER  & BPCER10 & BPCER20 & BPCER100 & D-EER  & BPCER10 & BPCER20 & BPCER100 \\ \midrule

ResNet34            & 4.44 & 2.22 & 4.44 & 16.67 & 8.89 & 8.89 & 12.22 & 20.00 & 4.44 & 3.33 & 3.33  & 5.56 & 5.56 & 4.44 & 5.56 & 20.00  \\     
ResNet101           & 4.44 & 2.22 & 4.44 & 6.67  & 5.56 & 4.44 & 4.44  & 8.89  & 4.44 & 0.00 & 3.33  & 4.44 & 4.44 & 3.33 & 4.44 & 7.78  \\ 
DenseNet121         & 6.67 & 3.33 & 7.78 & 18.89 & 6.67 & 5.56 & 6.67  & 14.44 & 3.33 & 2.22 & 2.22  & 10.00& 6.67 & 2.22 & 6.67 & 18.89  \\
MobileNetV3(L)      & 6.67 & 4.44 & 6.67 & 8.89  & 6.67 & 6.67 & 6.67  & 8.89  & 4.44 & 0.00 & 4.44  & 16.67& 6.67 & 5.56 & 6.67 & 15.56  \\
EfficientNetV2(S)   & 6.67 & 6.67 & 6.67 & 16.67 & 7.78 & 6.67 & 14.44 & 24.44 & 5.56 & 4.44 & 5.56  & 7.78 & 6.67 & 6.67 & 7.78 & 23.33  \\
\midrule 
                                        
Swin(Tiny)          & 5.56 & 0.00 & 4.44 & 15.56 & 7.78 & 4.44 & 23.33 & 48.89 & 4.44 & 1.11 & 4.44  & 28.89 & 5.93 & 1.11 & 8.89 & 38.89  \\     
Swin(Small)         & 2.22 & 0.00 & 0.00 & 5.56  & 3.33 & 0.00 & 0.00  & 6.67  & 2.22 & 0.00 & 0.00  & 3.33  & 2.41 & 0.00 & 0.00 &  5.56  \\ 
Swin(Base)          & 4.44 & 0.00 & 0.00 & 11.11 & 4.44 & 0.00 & 2.22  & 17.78 & 1.11 & 0.00 & 0.00  & 1.11 & 3.33 & 0.00 & 1.11 & 14.44  \\    \midrule 

CLIP(ViT-B-16)      & 2.22 & 0.00 & 0.00 & 4.44  & 3.33 & 0.00 & 2.22  & 4.44  & 2.22 & 0.00 & 0.00  & 2.22  &2.41 & 0.00 & 0.00  & 4.44  \\   
CLIP(ViT-B-32)      & 1.11 & 0.00 & 1.11 & 1.11  & 2.22 & 0.00 & 1.11  & 2.22  & 2.22 & 1.11 & 1.11  & 15.56  &2.22 & 1.11 & 1.11  & 3.33  \\ 
CLIP(ViT-L-14)      & 3.33 & 0.00 & 1.11 & 6.67  & 2.22 & 0.00 & 0.00  & 2.22  & 1.11 & 0.00 & 0.00  & 0.00 &2.22 & 0.00 & 0.00  & 4.44  \\   \midrule 

DINO(ViT-S-14)      & 2.22 & 0.00 & 0.00 & 4.44  & 4.44 & 0.00 & 0.00  & 13.33  & 1.11 & 0.00 & 0.00  & 0.00 & 2.41& 0.00 & 0.00  & 6.67  \\     
DINO(ViT-B-14)      & 1.11 & 1.11 & 1.11 & 1.11  & 3.33 & 1.11 & 2.22  & 24.44  & 2.22 & 1.11 & 1.11  & 3.33 & 2.22& 1.11 & 1.11  & 7.76\\
DINO(ViT-L-14)      & 1.11 & 0.00 & 0.00 & 0.00  & 2.22 & 0.00 & 2.22  &  3.33  & 1.11 & 0.00 & 0.00  & 1.11 & 2.22& 0.00 & 0.00  & 2.22   \\        
		\midrule
Avg.                & 3.73 & 1.51 & 2.70 & 8.41 & 4.92 & 2.70 & 5.55 & 14.28 & 2.86 & 0.95 & 1.82 & 7.14 & 3.96 & 1.83 & 3.10 & 12.38 \\ 
        \bottomrule
	 \end{tabular}
	\end{adjustbox}
\end{table*}

In line with the above goals, the experimental evaluation is carried out on five publicly available databases for PAD: CASIA-FASD~\cite{Zhang-CASIAFASD-ICB-2012} (denoted as C), REPLAY-ATTACK~\cite{Chingovska-REPLAYATTACK-BIOSIG-2012} (denoted as I), OULU-NPU~\cite{Boulkenafet-OULUNPU-FG-2017} (denoted as O), MSU-FASD~\cite{Wen-MSUFASD-TIFS-2015} (denoted as M), and SiW-Mv2~\cite{Guo-SiWMv2-ECCV-2022}. CASIA-FASD~\cite{Zhang-CASIAFASD-ICB-2012} database consists of 600 videos from 50 subjects, including warped-photo, cut-photo and video-replay attacks. REPLAY-ATTACK~\cite{Chingovska-REPLAYATTACK-BIOSIG-2012} contains 1,200 videos from 50 subjects and printed and replay attacks. OULU-NPU~\cite{Boulkenafet-OULUNPU-FG-2017} is a mobile facial PAD dataset, acquired with six different mobile phones and consisting of 4,950 videos from 55 subjects. MSU-FASD~\cite{Wen-MSUFASD-TIFS-2015} dataset includes printed photos and replay attacks, with a total of 440 videos from 35 subjects. SiW-Mv2~\cite{Guo-SiWMv2-ECCV-2022} is made up of 1,700 videos of 14 PAI species, including challenging attacks such as silicone masks, obfuscation and cosmetic make-up. Tab.~\ref{tab:DB} summarises the main characteristics of databases and Fig.~\ref{fig:databases} shows examples of BPs and PAIs for each dataset. In addition to the above scenarios, we also evaluated the four protocols defined in OULU-NPU~\cite{Boulkenafet-OULUNPU-FG-2017} that aim to assess the generalisability of PAD algorithms to unknown environmental conditions, unknown PAI species, interoperability of trapping devices and cross-database.  

\subsection{Implementation Details}

As the above databases contain videos, we followed~\cite{Fang-LMFD-WCACV-2022, Fang-CF-PAD-WCACV-2024} and sampled evenly 25 frames per video across the duration of each video. Subsequently, MTCNN~\cite{Zhang-MTCNN-2016} detects the face per frame, and the resulting image is resized to 256~$\times$~256 pixels. We also sampled the training data in each mini-batch as in~\cite{Shimizu-BalancedMiniBatch-AI4I-2018} to maintain a bona fide vs. attack ratio of 1:1. Additionally, face images are subjected to random data augmentation, e.g., change of the brightness, contrast, saturation and hue. While both traditional CNNs and DINO-based models were initialised with their pre-trained weights on ImageNet~\cite{Deng-ImageNet-CVPR0-2009}, the pre-trained weights of CLIP-based architectures stem from LAION-400M~\cite{Schuhmann-LAION400M-ArXiv-2021}. For DINO and CLIP, several backbones that divide the input image into different patch sizes and have a varying number of parameters (e.g. ViT-B-16, ViT-B-32 and ViT-L-14) were selected and evaluated. All algorithms were implemented in PyTorch~\cite{Paszke-PyTorchAnImperative-2019} and trained for 50 epochs using the Adam optimiser with a learning rate of $1\mathrm{e}{-4}$. A batch size of 128 images is set for training. In the inference phase, the final PAD score for a given video is computed as the fused score (mean-rule fusion) of all frames as done in~\cite{Fang-CF-PAD-WCACV-2024,Fang-LMFD-WCACV-2022,Liu-CIFAS-ICME-2022}.  

\begin{table*}[!t]
	\centering
	\caption{Detection performance (in \%) of foundation models for the SiW-Mv2 leave-one-out protocol. The best overall results are highlighted in bold.}
	\label{tab:SWIMv2}
	\begin{adjustbox}{width=\linewidth}
	\begin{tabular}{r|r| c c c c| c c c| c c c c c| c c| l}  \toprule
        \multirow{2}{*}{\textbf{Approaches}} & \multirow{2}{*}{\textbf{Metrics}}&\multicolumn{4}{c|}{\textbf{Covering}} & \multicolumn{3}{c|}{\textbf{Make-up}}& \multicolumn{5}{c|}{\textbf{3D Attack}} & \multicolumn{2}{c|}{\textbf{2D Attack}} &  \\
                                &         & FunE.& PEye& PMouth& PaperG.& Ob.& Impers.& Cosmetic& HalfM.& Silicone& TransM.& Paper & Mann.& Replay& Print & Avg.$\pm$Std. \\
            \midrule 
\multirow{2}{*}{SiWM-v2 baseline\cite{Guo-SiWMv2-ECCV-2022}} & BPCER100   &  91.10 & 63.00 & 11.60 & 96.00 & 1.70 & 76.20 & 60.80 & 38.60 & 52.50 & 0.00 & 0.00 & 33.4                                 & 60.70 & 21.10 & 43.34$\pm$33.19   \\
                                & HTER    & 29.50 & 2.70 & 1.10 & 11.90  & 1.30  & 24.50 & 10.90 & 8.00  & 9.20  & 0.00  & 0.60  & 4.00 & 17.90 & 9.60  & 9.40$\pm$8.80 \\ 
                                \midrule
\multirow{5}{*}{CLIP(ViT-B-16)} & D-EER   & 13.27 & 0.19 & 0.39  & 1.62  & 8.99  & 0.19  & 11.37 & 4.21  & 0.58  & 3.21  & 0.19  & 0.19 & 16.27 & 11.72 &   
                                5.17$\pm$5.85   \\
                                & BPCER10 & 14.67 & 0.39 & 0.39  & 0.39  & 8.11  & 0.39  & 15.44 & 2.70  & 0.39  & 2.32  & 0.00  & 0.39 & 17.76 & 15.44 & 5.63$\pm$7.04   \\
                                & BPCER20 & 33.20 & 0.39 & 0.77  & 1.16  &22.01  & 0.39  & 20.46 & 4.25  & 0.77  & 3.09  & 0.39  & 0.39 & 23.55 & 26.25 & 9.79$\pm$12.21   \\
                                & BPCER100& 50.19 & 0.39 & 0.77  & 1.93  &22.39  & 0.39  & 26.25 & 8.49  & 1.16  & 7.34  & 0.39  & 0.39 & 32.05 & 42.47 & 13.90$\pm$17.47   \\ 
                                & HTER    & 13.46 & 0.19 & 0.39  & 1.62  & 8.60  & 0.19  & 11.37 & 3.51  & 0.58  & 3.21  & 0.19  & 0.19 & 16.27 & 11.72 & 5.11$\pm$5.87   \\\midrule
\multirow{5}{*}{CLIP(ViT-B-32)} & D-EER   & 18.68 & 0.19 & 3.65  & 1.82  & 10.14 & 0.19  & 7.51  & 5.48  & 1.74  & 5.01  & 0.00  & 0.19 & 18.83 & 13.42 & 
                                6.20$\pm$6.67   \\
                                & BPCER10 & 33.98 & 0.39 & 1.16  & 0.39  & 11.20 & 0.39  & 4.63  & 3.86  & 0.39  & 1.93  & 0.00  & 0.39 & 32.05 & 16.22 & 7.64$\pm$11.75   \\
                                & BPCER20 & 45.95 & 0.39 & 3.86  & 0.39  & 15.83 & 0.39  & 12.74 & 5.41  & 3.47  & 5.02  & 0.00  & 0.39 & 37.84 & 20.85 & 10.90$\pm$14.71   \\
                                & BPCER100& 64.68 & 0.39 & 14.67 & 2.32  & 22.39 & 0.39  & 32.05 & 10.81 & 3.47  & 11.20 & 0.00  & 0.39 & 47.88 & 35.52 & 17.58$\pm$20.41   \\ 
                                & HTER    & 19.24 & 0.19 & 3.65  & 2.47  & 10.14 & 0.193 & 6.36  & 5.48  & 1.74  & 5.01 & 0.00 & 0.19   & 18.84 & 13.79 & 6.24$\pm$6.75   \\\midrule
\multirow{5}{*}{CLIP(ViT-L-14)} & D-EER   & 9.57  & 0.19 & 0.19  & 0.19  & 10.53 & 0.19  & 7.90  & 1.27  & 0.19  & 0.19  & 0.00  & 0.19 & 13.20 & 9.45  &   
                                \textbf{3.80$\pm$5.02}   \\
                                & BPCER10 & 9.27  & 0.39 & 0.39  & 0.39  & 11.97 & 0.39  & 8.11  & 0.39  & 0.39  & 0.39  & 0.00  & 0.38 & 15.06 & 8.49  & \textbf{4.00$\pm$5.34}   \\
                                & BPCER20 & 16.99 & 0.39 & 0.39  & 0.39  & 11.97 & 0.39  & 8.11  & 0.77  & 0.39  & 0.39  & 0.00  & 0.38 & 23.55 & 18.53 & \textbf{5.90$\pm$8.36}   \\
                                & BPCER100& 28.57 & 0.39 & 0.39  & 0.39  & 11.97 & 0.39  &23.94  & 1.16  & 0.39  & 0.39  & 0.00  & 0.38 & 41.31 & 37.84 & \textbf{10.54$\pm$15.49}   \\ 
                                & HTER    & 9.57  & 0.19 & 0.19  & 0.19  & 12.80 & 0.19  & 6.94  & 1.27  & 0.19  & 0.19  & 0.00 & 0.19  & 13.00 & 9.45  & \textbf{}\textbf{3.88$\pm$5.21}   \\\midrule
\multirow{5}{*}{DINO(ViT-S-14)} & D-EER   & 18.48 & 0.19 & 0.19  & 0.39  & 9.76  & 0.19  & 21.97 & 6.07  & 0.39  & 5.20  & 0.00  & 0.19 & 7.24  & 17.77 & 
                                6.29$\pm$7.84   \\
                                & BPCER10 & 37.07 & 0.39 & 0.39  & 0.39  & 10.42 & 0.39  & 30.89 & 4.25  & 0.39  & 3.86  & 0.00  & 0.39 & 6.18  & 35.14 & 9.30$\pm$3.96   \\
                                & BPCER20 & 44.40 & 0.39 & 0.39  & 0.39  & 13.51 & 0.39  & 35.52 & 6.56  & 0.39  & 5.41  & 0.00  & 0.39 & 8.49  & 57.53 & 12.41$\pm$19.04   \\
                                & BPCER100& 58.59 & 0.39 & 0.39  & 0.39  & 24.32 & 0.39  & 40.93 & 11.58 & 0.77  & 34.75 & 0.00  & 0.39 & 15.83 & 83.40 & 19.44$\pm$26.25   \\ 
                                & HTER    & 18.76 & 0.19 & 0.19  & 1.04  & 9.76  & 0.19  & 22.93 & 6.75  & 0.39  & 5.20  & 0.00  & 0.19 & 6.73  & 18.16 & 6.46$\pm$8.03   \\\midrule
\multirow{5}{*}{DINO(ViT-B-14)} & D-EER   & 7.96  & 0.19 & 0.19  & 0.19  & 18.16 & 0.39  & 19.07 & 1.47  & 0.60  &  1.61 & 0.00  & 0.19 & 6.54  & 13.62 & 
                                5.01$\pm$7.01   \\
                                & BPCER10 & 6.95  & 0.39 & 0.39  & 0.39  & 27.03 & 0.39  & 32.82 & 0.77  & 0.77  &  0.39 & 0.00  & 0.39 & 3.86  & 30.12 & 7.48$\pm$12.40   \\
                                & BPCER20 & 11.20 & 0.39 & 0.39  & 0.39  & 30.12 & 0.39  & 42.08 & 0.77  & 0.77  &  0.77 & 0.00  & 0.39 & 9.27  & 49.03 & 10.43$\pm$17.04   \\
                                & BPCER100& 25.87 & 0.39 & 0.39  & 0.39  & 61.00 & 0.39  & 68.34 & 1.54  & 1.16  &  1.16 & 0.00  & 0.39 & 21.62 & 64.09 & 17.62$\pm$26.74  \\ 
                                & HTER    & 8.52  & 0.19 & 0.19 & 0.19   & 18.16 & 0.39  & 18.88 & 1.47  & 0.58  & 1.41  & 0.00  & 0.19 & 7.046 & 13.99 & 5.09$\pm$7.05   \\\midrule
\multirow{5}{*}{DINO(ViT-L-14)} & D-EER   & 21.90 & 0.19 & 0.19 & 0.19   & 9.95  & 0.19  & 20.04 & 0.39  & 0.77  &  0.19 &  0.00 & 0.19 & 8.14  & 11.72 & 
                                5.29$\pm$7.81   \\
                                & BPCER10 & 53.67 & 0.39 & 0.39 & 0.39   &10.81  & 0.39  & 36.29 & 0.39  & 0.39  &  0.39 &  0.00 & 0.39 & 6.56  & 21.62 & 9.43$\pm$16.59   \\
                                & BPCER20 & 64.86 & 0.39 & 0.39 & 0.39   &10.81  & 0.39  & 42.47 & 0.39  & 0.77  &  0.39 &  0.00 & 0.39 & 11.97 & 42.86 & 12.61$\pm$21.28   \\
                                & BPCER100& 90.35 & 0.39 & 0.39 & 0.39   &15.44  & 0.39  & 57.14 & 0.39  & 1.54  &  0.39 &  0.00 & 0.39 & 32.43 & 78.38 &  19.86$\pm$32.04  \\
                                & HTER    & 21.62 & 0.19 & 0.19 & 0.19   & 7.68  & 0.19  & 21.00 & 1.08  & 0.77  &  0.19 &  0.00 & 0.19 &  7.56 & 11.72 & 5.18$\pm$7.78   \\
\bottomrule
	 \end{tabular}
	\end{adjustbox}
\end{table*}

\subsection{Evaluation Metrics}

The experimental results are analysed and reported in compliance with the metrics defined in the international standard ISO/IEC 30107-3~\cite{ISO-IEC-30107-3-PAD-metrics-2023} for biometric PAD:

\begin{itemize}
    \item Attack Presentation Classification Error Rate (APCER), which computes the proportion of attack presentations wrongly classified as bona fide presentations.

    \item Bona Fide Presentation Classification Error Rate (BPCER), which is defined as the proportion of bona fide presentations misclassified as attack presentations.
\end{itemize}

Based on these metrics, we report $i)$ the BPCERs observed at APCER values or security thresholds of 1\% (BPCER100), 5\% (BPCER20), and 10\% (BPCER10); and $ii)$ the Detection Equal Error Rate (D-EER), which is defined as the error rate value at the operating point where APCER = BPCER. To benchmark against the state of the art, non-ISO compliant metrics are also presented, i.e., Half-Total Error Rate (HTER) and Area Under the Receiver Operating Characteristic (ROC) Curve (AUC).

\section{RESULTS AND DISCUSSION}
\label{sec:results}

Following the above goals, the next sections present the effectiveness evaluation of foundation models in different operational scenarios defined in Sect.~\ref{sec:exp_setup} (i.e., known-attacks~\ref{sec:known_attacks}, unknown PAI species~\ref{sec:unknown_attacks} and cross-database~\ref{sec:cross_db}). The foundation model-based PAD framework is also evaluated for the most challenging scenario in Sect.~\ref{sec:cross_db}.  

\subsection{Known-Attacks}
\label{sec:known_attacks}

Tab.~\ref{tab:KA_CASIA} reports the detection performance of foundation models for zero-shot PAD in the simple CASIA-FASD known-attack scenario and benchmarks them against traditional CNNs. Note that both DINO and CLIP yield overall D-EERs (last columns) of less than 2.41\%, together with BPCERs between 2.22\% and 7.76\% for high-security thresholds (i.e. APCER=1\%). Compared to the results collected by traditional CNNs (e.g. ResNet34, DenseNet121 and EfficientNetV2(S)), the detection performances of DINO and CLIP are up to 10 times lower for the same security threshold (BPCER100), demonstrating their soundness in terms of generalisability. We can also observe a significant improvement in the performance of the foundation models with respect to networks based on vision transformers (i.e. Swin~\cite{Liu-SwinTrans-ICCV-2021}), even though the latter are the basis of the foundation models. In particular, the Swin models yield an overall BPCER100 in the ranges 5.56\%-38.89\%, which are considerably higher than those obtained by DINO (BPCER100~$\leq$~7.76\%) and CLIP (BPCER100~$\leq$~4.44\%). These unreliable detection results of Swin architectures for higher security thresholds indicate that the model will significantly reduce its performance for more challenging scenarios and are therefore discarded for further analysis. Notice that a comparison between the two foundation models in this scenario is not feasible, as their overall performance is similar in terms of D-EER and is statistically approximated for higher security thresholds (i.e. mean BPCER100~(CLIP) of 4.07\% vs. mean BPCER100~(DINO) of 5.55\%).   

It should be noted that the reported results for the different attacks vary depending on the PAI species, with the video-replay attack being on average the most difficult to detect. The BPCER100 value for the video-replay attack is, on average, almost twice as high as that recorded for other PAI species (14.28\% vs. 8.41\% - cut-photo vs. 7.14\%- warped-photo). This indicates that the artefacts or attack traces produced by the video replay against the biometric capture device are partially encoded by the deep neural networks. The latter trend is different between DINO and CLIP. While DINO performs on average worse for video-replay attacks, CLIP does worse for warped-photo attacks. Therefore, we strongly believe that a score-level fusion between the two foundation models through the zero-shot PAD framework presented in Sect.~\ref{sec:framework} could benefit the final decision - the detection performance improvement by the fusion can be observed in Sect.~\ref{sec:cross_db}.  

\subsection{Unknown PAI species}
\label{sec:unknown_attacks}

We evaluate the generalisability of the foundation models for the challenging scenario of unknown PAI species, including 3D masks (i.e. silicone masks, transparent masks and mannequin head) and make-up (obfuscation, impersonation and cosmetic). For this purpose, the SiW-Mv2~\cite{Guo-SiWMv2-ECCV-2022} database is used and the leave-one-out protocol is followed: thirteen PAI species are used for training and the remaining PAI species is tested. Tab.~\ref{tab:SWIMv2} reports in compliance with ISO/IEC 30107-3 and benchmarks against the SiW-Mv2 baseline in terms of HTER and BPCER100. Note that all foundation models significantly outperform the reference model, reducing the latter's HTER = 9.40\% down to 3.88\% and its BPCER100 = 43.34\% down to 10.54\%. While the baseline PAD model rejects almost half of 100 bona fide presentation transactions when the system threshold is set to APCER = 1\%, the CLIP(VIT-L-14) model only rejects at most 10 out of 100 BP samples for the same threshold. Observe also that most of the foundation models achieve lower error rates for the challenging 3D attacks. In particular, CLIP(VIT-L-14) reports HTERs in the ranges 0\% to 1.27\%, which are significantly lower than those recorded by the baseline (HTER of up to 8.00\% for Half Masks). Similar trends can be observed for a high-security threshold: the baseline subsystem achieves a BPCER100 = 52.50\% for the silicone mask, while CLIP(VIT-L-14) reduces it down to 0.39\%.   
        
\begin{figure}[!b]
    \centering 
    \begin{subfigure}{0.48\linewidth}
        \includegraphics[width=\linewidth]{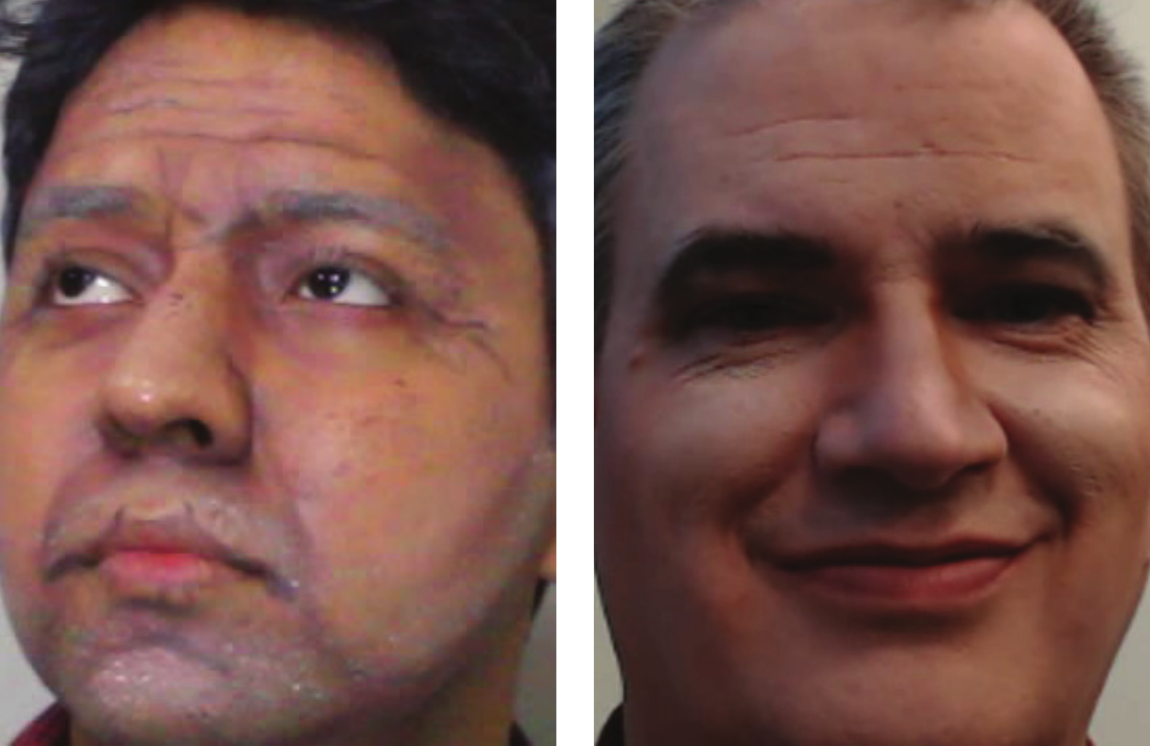}
        \caption{Make-up}
    \end{subfigure}
    \begin{subfigure}{0.48\linewidth}
        \includegraphics[width=\linewidth]{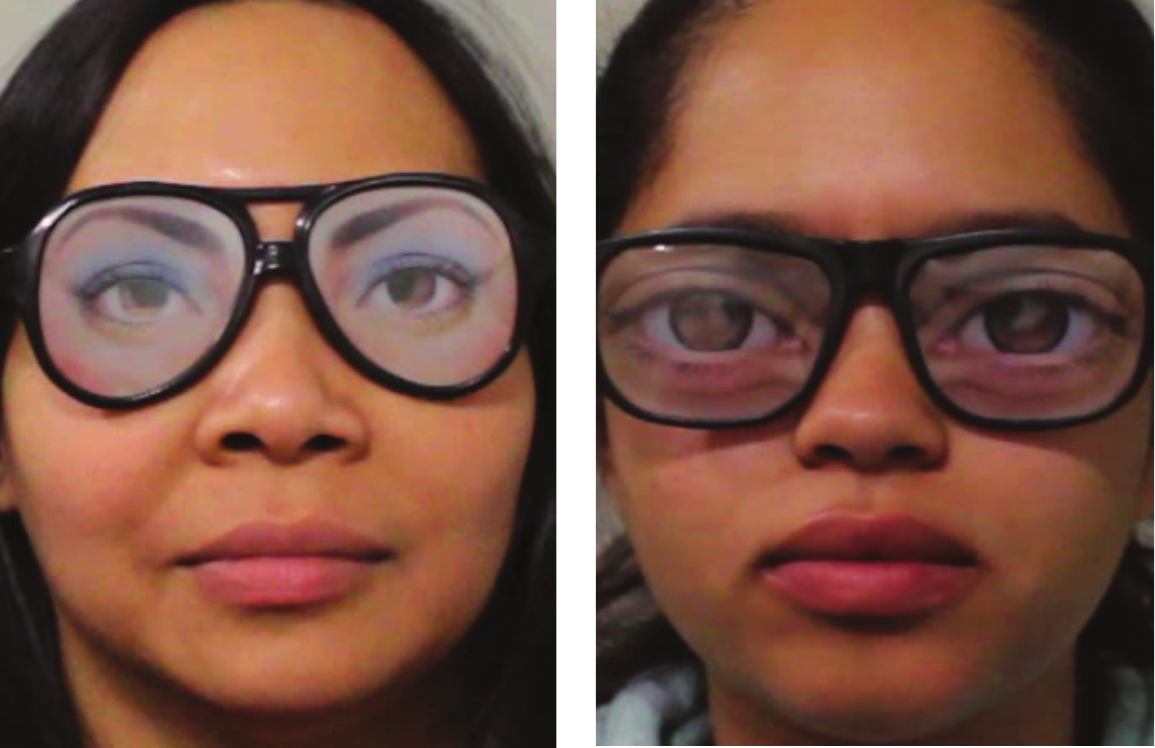}
        \caption{Funny-eyes}
    \end{subfigure}
    \caption{Example of challenging PAI species.}
    \label{fig:pai_species}
\end{figure}

\begin{table*}[!t]
	\centering
	\caption{Benchmark (in \%) of foundation models against the state of the art for different cross-database settings. The best results are highlighted in bold.}
	\label{tab:CROSSDB}
	\begin{adjustbox}{width=0.8\linewidth}
	\begin{tabular}{r| c c| c c| c c| c c| c c}  \toprule
\multirow{2}{*}{\textbf{Approaches}} & \multicolumn{2}{c|}{\textbf{O\&C\&I $\rightarrow$ M}} & \multicolumn{2}{c|}{\textbf{O\&M\&I $\rightarrow$ C}}  & \multicolumn{2}{c|}{\textbf{O\&C\&M $\rightarrow$ I}} & \multicolumn{2}{c|}{\textbf{I\&C\&M $\rightarrow$ O}} & \multicolumn{2}{c}{\textbf{Avg.}} \\ 
                                &   HTER$\downarrow$ & AUC$\uparrow$ &  HTER$\downarrow$ & AUC$\uparrow$ & HTER$\downarrow$ & AUC$\uparrow$  &  HTER$\downarrow$ & AUC$\uparrow$ &  HTER$\downarrow$ & AUC$\uparrow$ \\
        \midrule 
MADDG~\cite{Shao-MADDG-CVPR-2019}        &  17.69 &  88.06  &   24.50   &  84.51   &   22.19    &   84.99   &   27.89 &  80.02  &   23.07   &   84.40   \\
RFM~\cite{Shao-RFM-AI-2020}              & 17.30  &  90.48  &   13.89   &  93.98   &   20.27    &   88.16   &   16.45 &  91.16  &   16.98   &   90.95   \\
SSDG-R~\cite{Jia-SSDG-CVPR-2020}         &  7.38  &  97.17  &   10.44   &  95.94   &   11.71    &   96.59   &   15.61 &  91.54  &   11.29   &   95.31  \\
D$^2$AM~\cite{Chen-D2AM-AI-2021}         & 12.70  &  95.66  &   20.98   &  85.58   &   15.43    &   91.22   &   15.27 &  90.87  &   16.10   &   90.83   \\
ViT~\cite{Huang-AdaTransVit-ECCV-2022}   &  \textbf{4.75}  &  \textbf{98.59}  &   15.70   &  92.76   &    17.68   &   86.66   &   16.46 &  90.37  &   13.65   &   92.10   \\
TransFAS~\cite{Wang-TransFas-TBIOM-2022} &  7.08  &  96.69  &   9.81    &  96.13   &    10.12   &   95.53   &   15.52 &  91.10  &   10.63   &   94.86   \\
LMFD-PAD~\cite{Fang-LMFD-WCACV-2022}     &  10.48  &  94.55  &   12.50  &  94.17   &    18.49   &   84.72   &   13.47 &  92.09  &   10.63   &   94.86   \\
DADN-CDS~\cite{Yan-DADNcDS-2022}         &  5.24  &  98.06  &   6.84    &  97.95   &    10.64   &   95.14   &   13.77 &  93.09  &   \textbf{9.12}    &   96.06     \\
CIFAS~\cite{Liu-CIFAS-ICME-2022}         &  5.95  &  96.32  &   10.66   &  95.30   &    \textbf{8.50}    &   \textbf{97.24}   &   13.17 &  93.44  &   9.57    &   95.58     \\
CF-PAD~\cite{Fang-CF-PAD-WCACV-2024}     &  8.11  &  96.43  &   11.78   &  95.64   &   16.50    &   91.50   &   9.87  &  95.13  &   11.57   &   94.68     \\
MDIL~\cite{Wang-MDIL-AI-2024}            &  5.71  &  98.19  &   13.22   &  91.94   &   11.25    &   95.44   &   12.47 &  94.22  &   10.66   &   94.95     \\
FoundPAD (Vit-B)~\cite{Ozgur-FoundPAD-ArXiv-2025} &  20.95  &  89.88   &  \textbf{4.89}    &   98.08    &   10.45   &   95.80 &   6.19  &   98.31   &   10.62 &   95.52 \\
FoundPAD (Vit-L)~\cite{Ozgur-FoundPAD-ArXiv-2025} &  16.90  &  93.18   &  6.00    &   \textbf{98.72}    &    9.90   &   96.07 &   \textbf{5.87}  &   \textbf{98.41}   &   9.67  &   \textbf{96.60} \\
\midrule
CLIP(ViT-B-16)                           &  21.58 & 86.20   &   12.59   & 94.79    &    32.42   &   70.37   &   27.17 & 80.97   &    23.44  &   83.08  \\
CLIP(ViT-B-32)                           &  21.94 & 86.55   &   25.37   & 84.99    &    26.31   &   79.62   &   26.36 & 82.08   &    25.00  &   83.31  \\
CLIP(ViT-L-14)                           &  23.38 & 85.40   &   12.04   & 95.51    &    28.28   &   77.10   &   22.34 & 85.61   &    21.51  &   85.91  \\
DINO(ViT-S-14)                           &  21.94 & 89.40   &   22.41   & 84.25    &    22.63   &   82.96   &   31.14 & 76.26   &    24.53  &   83.22  \\    
DINO(ViT-B-14)                           &  21.58 & 85.50   &   15.37   & 94.02    &    21.08   &   88.44   &   25.58 & 82.64   &    20.90  &   87.65   \\
DINO(ViT-L-14)                           &  20.14 & 87.44   &   14.44   & 94.16    &    14.80   &   92.85   &   15.86 & 92.30   &    16.31  &   91.69   \\
\midrule
$\mathbf{MAX}$[DINO(ViT-L-14), CLIP(ViT-L-14)] &  21.23 & 85.61  &  7.41 & 96.53 & 15.07 & 91.11  & 14.32 & 91.74 &  14.51 & 91.25  \\
$\mathbf{MIN}$[DINO(ViT-L-14), CLIP(ViT-L-14)] &  17.27 & 91.63  &  9.07 & 96.95 & 16.68 & 90.63  & 17.08 & 90.97 &  15.03 & 92.55 \\
$\mathbf{SUM}$[DINO(ViT-L-14), CLIP(ViT-L-14)] &  17.27 & 90.63  &  5.93 & 97.60 & 14.04 & 91.89  & 15.34 & 92.35 &  13.15 & 93.12 \\
$\mathbf{AVG}$[DINO(ViT-L-14), CLIP(ViT-L-14)] &  17.27 & 90.63  &  5.93 & 97.60 & 14.04 & 91.89  & 15.34 & 92.35 &  13.15 & 93.12  \\
	\bottomrule
	 \end{tabular}
	\end{adjustbox}
\end{table*}

Taking a closer look at Tab.~\ref{tab:SWIMv2}, we can also note that most PAD techniques have poor detection performance for funny-eyes and cosmetic attacks: D-EERs are close to 15\%, which makes them the most difficult PAI species. This is because the make-up applied to the faces is subtle, and therefore they look like real human faces. Funny-eye attacks contain a part of the face image that belongs to bona fide users, which makes it difficult for PAD subsystems to detect (see Fig.~\ref{fig:pai_species}). Patch-centric classification could be a potential solution to improve detection performance on this latter attack. A proper assessment reporting the impact of funny-eyes attacks on the real face recognition system can show whether such attacks pose a real threat and, thus, whether they could lead to a false match. 

\subsection{Cross-database}
\label{sec:cross_db}

\begin{figure*}[!tb]
    \centering
    \begin{subfigure}{0.32\linewidth}
        \includegraphics[width=\linewidth]{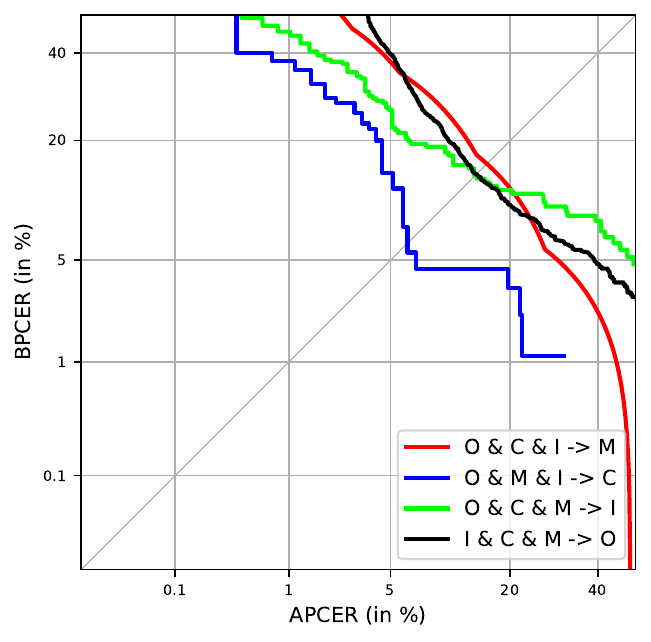}
        \caption{$\mathbf{AVG}$}
    \end{subfigure}
    \begin{subfigure}{0.32\linewidth}
        \includegraphics[width=\linewidth]{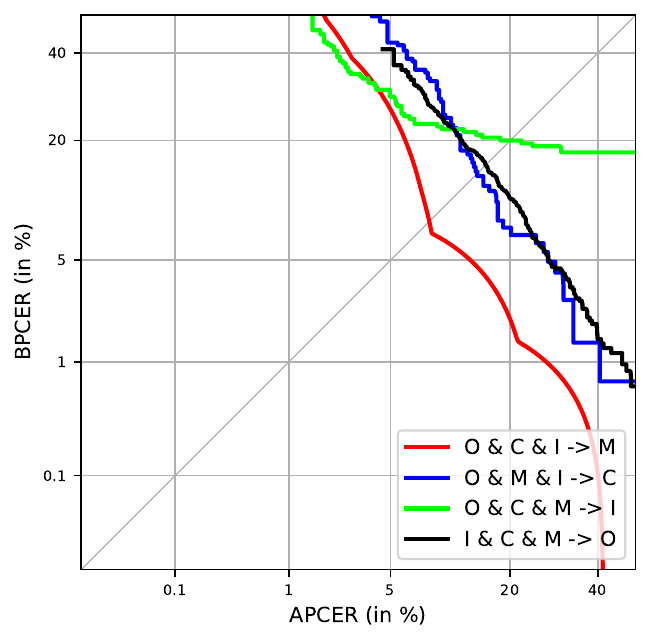}
        \caption{CF-PAD~\cite{Fang-CF-PAD-WCACV-2024}}
    \end{subfigure}
    \begin{subfigure}{0.32\linewidth}
        \includegraphics[width=\linewidth]{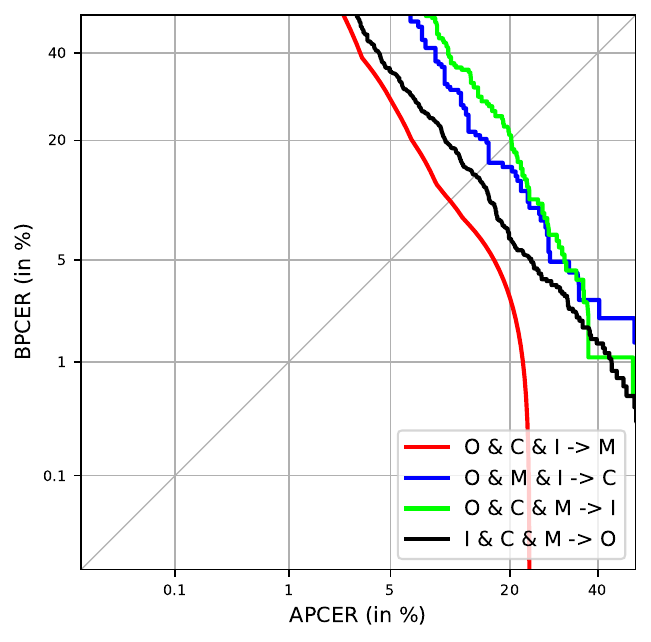}
        \caption{LMFD-PAD~\cite{Fang-LMFD-WCACV-2022}}
    \end{subfigure}
    \caption{In-depth performance benchmark of our average fusion (i.e., $\mathbf{AVG}$[DINO(ViT-L-14), CLIP(ViT-L-14)]) with two state-of-the-art high-performance PAD approaches in terms of BPCER vs. APCER.}
    \label{fig:cross_db}
\end{figure*}

The development of PAD subsystems has evolved rapidly over the years, especially with the introduction of deep neural networks. Contrary to technological progress, the creation of new databases to train and achieve the generalisability of such algorithms is slower due to certain privacy issues and is a time-consuming task. In real applications, the phenomenon of data drift, which includes changes in environmental conditions, unknown PAI species and even subject changes, leads to a shift in the statistical distribution of test images and thus to poor PAD performance. In Tab.~\ref{tab:CROSSDB}, the generalisability of foundation models for zero-shot PADs in cross-database scenarios where data drift exists is reported. Following previous works~\cite{Liu-CIFAS-ICME-2022,Fang-CF-PAD-WCACV-2024,Wang-MDIL-AI-2024,Ozgur-FoundPAD-ArXiv-2025}, we perform four training-test configurations, i.e., O\&C\&I~$\rightarrow$~M, O\&M\&I~$\rightarrow$~C, O\&C\&M~$\rightarrow$~I, and I\&C\&M~$\rightarrow$~O. Note that foundation models (both DINO and CLIP) achieve on average the state-of-the-art performance. While the most advanced methods specifically designed for PAD yield HTERs between 9.12\% and 23.07\%, HTERs from general-purpose foundation models range on average between 16.31\% and 25.0\%, considering only the optimisation of the classification header. 

To find out to what extent foundation models can improve zero-shot PAD, we also report on the score-level fusion PAD framework ($\mathbf{MAX}$, $\mathbf{MIN}$, $\mathbf{SUM}$, $\mathbf{AVG}$) between the best-performing model per category (i.e. DINO(ViT-L-14) and CLIP(ViT-L-14)) in Tab.~\ref{tab:CROSSDB}. Note that all score-level mergers can be carried out without the need to be adjusted over a development set. We observe that both the average ($\mathbf{AVG}$) and the sum ($\mathbf{SUM}$) between the scores computed by DINO and CLIP result in a significant improvement of the detection performance of each foundation model separately. In particular, $\mathbf{AVG}$[DINO(ViT-L-14), CLIP(ViT-L-14)] computes a HTER and AUC of 13.15\% and 93.12\%, respectively, which are even closer to the state-of-the-art performance. We strongly believe that more sophisticated score-level fusions (e.g. boosting, bagging and weighted voting) could further improve past performance.

\subsubsection{In-depth Performance Analysis}

Since both HTER and AUC are not ISO/IEC-compliant metrics and are not completely reliable for measuring the algorithm performance (they oversimplify the trade-off between APCER and BPCER and are threshold sensitive), we compare our $\mathbf{AVG}$[DINO(ViT-L-14), CLIP(ViT-L-14)]) against two of the state-of-the-art approaches in terms of APCER vs. BPCER in Fig.~\ref{fig:cross_db}. To plot the DET curves of CF-PAD~\cite{Fang-CF-PAD-WCACV-2024} and LMDF-PAD~\cite{Fang-LMFD-WCACV-2022},  we used their pre-trained weights and pre-processed the images as in their respective articles. Therefore, the performance shown by them in Fig.~\ref{fig:cross_db} may differ slightly from that reported in Tab.~\ref{tab:CROSSDB}. 

Despite our $\mathbf{AVG}$[DINO(ViT-L-14), CLIP(ViT-L-14)]) was only adjusted for zero-shot PAD, we can observe in Fig.~\ref{fig:cross_db} that it outperforms the state-of-the-art for most security operational thresholds: the BPCER@APCER=1\% of our zero-shot PAD framework is lower than the one yielded by CF-PAD~\cite{Fang-CF-PAD-WCACV-2024} and LMDF-PAD~\cite{Fang-LMFD-WCACV-2022}, respectively. Note that CF-PAD~\cite{Fang-CF-PAD-WCACV-2024} is a domain adaptation approach specifically designed for cross-database scenarios where source and target domains differ. Conceptually, domain adaptation and zero-shot learning deal, in different ways, with the phenomenon of data drift. However, the results show that zero-shot learning through foundation models opens up, with minimal effort, a new avenue for addressing PAD. The results also indicate, on the one hand, that the current comparison assessment in terms of HTER and AUC is not fully reliable and should be replaced by BPCER values in different APCERs in future benchmarking. On the other hand, pre-trained weights from foundation models can be widely used for zero-shot PAD and can, therefore, be combined with previous PAD approaches~\cite{Fang-PatchWise-IJCB-2022} using traditional networks to improve their performance.        

\subsubsection{Further Generalisability Analysis}

\begin{table}[!t]
	\centering
	\caption{Detection performance (in \%) of foundation models for different OULU-NPU protocols. The best results are highlighted in bold.}
	\label{tab:OULU}
	\begin{adjustbox}{width=\linewidth}
	\begin{tabular}{c| r| c c c}  \toprule
        \textbf{P}              & \textbf{Approaches} & \textbf{HTER}  & \textbf{APCER} & \textbf{BPCER} \\ \midrule                                   
\multirow{11}{*}{\textbf{1}}    &  LMFD-PAD~\cite{Fang-LMFD-WCACV-2022}               &    1.50        & 1.40 & 1.60  \\
                                &  PatchSwap~\cite{Fang-PatchWise-IJCB-2022}          &    0.60        & 0.40 & 0.80 \\
                                &  CDCN++~\cite{Yu-CDCNPlusPlus-CVPR-2020}            &    0.20        & 0.40 & \textbf{0.00} \\
                                &  NAS-FAS~\cite{Yu-NASfAS-PAMI-2020}                 &    0.20        & 0.40 & \textbf{0.00} \\
                                &  PatchNet~\cite{Wang-PatchNet-CVPR-2022}            &    \textbf{0.00} & \textbf{0.00} & \textbf{0.00} \\ \cmidrule{2-5}
                                & $\mathbf{MAX}$[DINO(ViT-L-14), CLIP(ViT-L-14)]      &    2.64        &    2.50    &    2.77    \\
                                & $\mathbf{MIN}$[DINO(ViT-L-14), CLIP(ViT-L-14)]      &    5.35        &    5.69    &    5.00    \\
                                & $\mathbf{SUM}$[DINO(ViT-L-14), CLIP(ViT-L-14)]      &    4.65        &    4.31    &    5.00    \\
                                & $\mathbf{AVG}$[DINO(ViT-L-14), CLIP(ViT-L-14)]      &    4.65        &    4.31    &    5.00     \\
\midrule
\multirow{11}{*}{\textbf{2}}    &  LMFD-PAD~\cite{Fang-LMFD-WCACV-2022}               &    2.00        &   3.10      &  \textbf{0.80}  \\
                                &  PatchSwap~\cite{Fang-PatchWise-IJCB-2022}          &    1.80        &   2.50      &  1.10  \\
                                &  CDCN++~\cite{Yu-CDCNPlusPlus-CVPR-2020}            &    1.30        &   1.80      &  \textbf{0.80}  \\
                                &  NAS-FAS~\cite{Yu-NASfAS-PAMI-2020}                 & \textbf{1.20}  &   1.50      &  \textbf{0.80}  \\
                                &  PatchNet~\cite{Wang-PatchNet-CVPR-2022}            & \textbf{1.20}  &\textbf{1.10}&  1.20  \\ \cmidrule{2-5}
                                & $\mathbf{MAX}$[DINO(ViT-L-14), CLIP(ViT-L-14)]      &    2.13        &   2.04      &  2.22    \\
                                & $\mathbf{MIN}$[DINO(ViT-L-14), CLIP(ViT-L-14)]      &    4.44        &   4.44      &  4.44     \\
                                & $\mathbf{SUM}$[DINO(ViT-L-14), CLIP(ViT-L-14)]      &    3.24        &   3.15      &  3.33      \\
                                & $\mathbf{AVG}$[DINO(ViT-L-14), CLIP(ViT-L-14)]      &    3.24        &   3.15      &  3.33    \\
\midrule
\multirow{11}{*}{\textbf{3}}    &  LMFD-PAD~\cite{Fang-LMFD-WCACV-2022}               & 3.40$\pm$3.10 & 3.50$\pm$3.20 & 3.40$\pm$3.10  \\
                                &  PatchSwap~\cite{Fang-PatchWise-IJCB-2022}          & 3.30$\pm$4.90 & \textbf{1.40$\pm$1.30} & 5.30$\pm$10.00  \\
                                &  CDCN++~\cite{Yu-CDCNPlusPlus-CVPR-2020}            & 1.80$\pm$0.70 & 1.70$\pm$1.50 & 2.00$\pm$1.20  \\
                                &  NAS-FAS~\cite{Yu-NASfAS-PAMI-2020}                 & 1.70$\pm$0.60 & 2.10$\pm$1.30 & 1.40$\pm$1.10  \\
                                &  PatchNet~\cite{Wang-PatchNet-CVPR-2022}            & \textbf{1.20$\pm$1.30} & 1.80$\pm$1.47 & \textbf{0.56$\pm$1.24}\\ \cmidrule{2-5}
                                & $\mathbf{MAX}$[DINO(ViT-L-14), CLIP(ViT-L-14)]      & 2.61$\pm$0.66 & 2.63$\pm$0.67 & 2.59$\pm$0.65 \\
                                & $\mathbf{MIN}$[DINO(ViT-L-14), CLIP(ViT-L-14)]      & 2.72$\pm$0.52 & 2.70$\pm$0.61 & 2.74$\pm$0.44 \\
                                & $\mathbf{SUM}$[DINO(ViT-L-14), CLIP(ViT-L-14)]      & 2.06$\pm$0.35 & 2.06$\pm$0.35 & 2.07$\pm$0.36  \\
                                & $\mathbf{AVG}$[DINO(ViT-L-14), CLIP(ViT-L-14)]      & 2.06$\pm$0.35 & 2.06$\pm$0.35 & 2.07$\pm$0.36 \\
\midrule
\multirow{11}{*}{\textbf{4}}    &  LMFD-PAD~\cite{Fang-LMFD-WCACV-2022}           & 3.30$\pm$3.10 & 2.50$\pm$4.10 & 3.30$\pm$3.10 \\
                                &  PatchSwap~\cite{Fang-PatchWise-IJCB-2022}      & 3.80$\pm$6.30 & 2.50$\pm$8.30 & 5.00$\pm$4.20  \\
                                &  CDCN++~\cite{Yu-CDCNPlusPlus-CVPR-2020}        & 5.00$\pm$2.90 & 4.20$\pm$3.40 & 5.80$\pm$4.90  \\
                                &  NAS-FAS~\cite{Yu-NASfAS-PAMI-2020}             & \textbf{2.90$\pm$2.80} & \textbf{2.10$\pm$1.30} & \textbf{1.40$\pm$1.10}  \\
                                &  PatchNet~\cite{Wang-PatchNet-CVPR-2022}        & 2.90$\pm$3.00 & 2.50$\pm$3.81 & 3.33$\pm$3.73  \\   \cmidrule{2-5} 
                                & $\mathbf{MAX}$[DINO(ViT-L-14), CLIP(ViT-L-14)]  & 3.42$\pm$0.85 & 3.06$\pm$1.12 &  3.78$\pm$0.69 \\
                                & $\mathbf{MIN}$[DINO(ViT-L-14), CLIP(ViT-L-14)]  & 6.53$\pm$1.21 & 6.50$\pm$1.22 &  6.56$\pm$1.22 \\
                                & $\mathbf{SUM}$[DINO(ViT-L-14), CLIP(ViT-L-14)]  & 5.14$\pm$1.71 & 5.17$\pm$1.88 &  5.11$\pm$1.56 \\
                                & $\mathbf{AVG}$[DINO(ViT-L-14), CLIP(ViT-L-14)]  & 5.14$\pm$1.71 & 5.17$\pm$1.88 &  5.11$\pm$1.56 \\
		\bottomrule
	 \end{tabular}
	\end{adjustbox}
\end{table}

Tab.~\ref{tab:OULU} benchmark also the foundation models framework against the state-of-the-art on OULU-NPU~\cite{Boulkenafet-OULUNPU-FG-2017}. Similar to the results in Tab.~\ref{tab:CROSSDB}, we observe that the score-level fusion between the two best-performing zero-shot foundation models reaches the state of the art in most of the OULU-NPU protocols. While the most advanced PAD methods return HTER values ranging from 0\% to 5\%, our zero-shot approaches return values between 2\% and 6\% for the same metric. Based on the trends in Tab.~\ref{tab:OULU}, we believe that an in-depth analysis of the performance of the foundation models and the state of the art for different security thresholds on OULU-NPU may show similar trends to those in Fig.~\ref{fig:cross_db}.

\section{CONCLUSIONS AND FUTURE WORKS}
\label{sec:conclusions}

In this work, we conducted an in-depth analysis of the best-performing foundation models for zero-shot PAD, which demonstrated the potential of these models to achieve generalisable classification even with low data availability. For this purpose, the pre-trained DINO and CLIP foundation models were selected and their classification header modified to a single neuron, only optimised to produce a zero-shot classification. This enabled us to find out whether their pre-trained weights optimised during a self-supervised training process were sufficiently generalisable to deliver detection performance close to the state of the art in challenging unknown scenarios. In the experimental evaluation of well-established databases and protocols, we evaluated the combination of different backbones having varying numbers of parameters, together with both foundation models. Experimental results show that DINO and CLIP can obtain detection results close to or even superior to those produced by the state-of-the-art methods, which were specifically designed for PAD. 

We also proposed a simple and effective zero-shot PAD framework that performs a score-level fusion between the best-performing backbone (i.e. VIT-L-14) of DINO and CLIP. The results showed that simple fusion strategies are beneficial for zero-shot PAD, resulting in a significant improvement of the base models in the most difficult scenario (i.e. cross-database). Regarding the latter, in-depth performance analysis in terms of DET curves (see Fig.~\ref{fig:cross_db}) revealed that non-ISO/IEC compliant metrics such as HTER and AUC in Tab.~\ref{tab:CROSSDB} are not fully reliable: our zero-shot fusion framework outperformed the two high-performing PAD approaches for high-security thresholds: these operating points are of the utmost importance to the industry during the deployment of PAD algorithms in real-world applications. We believe, on the one hand, that more sophisticated score-level fusions (e.g. boosting, bagging and weighted voting) could further improve the detection performance. On the other hand, our work demonstrated that pre-trained weights from foundation models can be widely used for PAD and can therefore be combined with previous PAD approaches~\cite{Fang-PatchWise-IJCB-2022} using traditional networks to improve their performance. 

For the future, we plan to leverage the text prompt to inject, during inference, additional knowledge extracted from the faces into the foundation models to further improve their detection performance.   

\section*{ETHICAL IMPACT STATEMENT}
This research complies with all ethical guidelines established by Face and Gesture 2025. The dataset was collected following the recommendations of the providers. All data has been anonymised to ensure that no individual can be discriminated against on the basis of gender, ethnicity or any other characteristic. In addition, all datasets have been previously used in different publications and competitions.


{\small
\bibliographystyle{ieee}
\bibliography{egbib}
}

\end{document}